# Unsupervised spectral learning

**Susan Shortreed**
Department of Statistics
University of Washington
Seattle, WA 98195

**Marina Meilă**
Department of Statistics
University of Washington
Seattle, WA 98195

## Abstract

In spectral clustering and spectral image segmentation, the data is partitioned starting from a given matrix of pairwise similarities S. The matrix S is constructed by hand, or learned on a separate training set. In this paper we show how to achieve spectral clustering in unsupervised mode. Our algorithm starts with a set of observed pairwise features, which are possible components of an unknown, parametric similarity function. This function is learned iteratively, at the same time as the clustering of the data. The algorithm shows promising results on synthetic and real data.

## 1 Introduction

While there has been much progress in obtaining better spectral clusterings with similarities given or constructed by hand, the problem of automatically estimating the similarities from data has received less attention. This limits the application of spectral clustering to the ability of the domain experts to guess the correct features and their optimal combination for each problem. It makes the results of the clustering algorithm sensitive to the particular function chosen. To overcome this limitation, several methods for learning the similarity function using a training set of clustered points have been proposed [Meilă and Shi, 2001a, Bach and Jordan, 2004, Meilă et al., 2005, Cour et al., 2005].

In the present paper, we take the unsupervised approach to spectral learning. We assume that we are given a set of points $\mathcal{D}$, several pairwise features measured on $\mathcal{D}$, and a similarity function $S$ that depends on the features through some parameters $\theta$. The algorithm we introduce adapts the parameter values in order to obtain a good clustering on $\mathcal{D}$ in a manner reminiscent of the alternating minimization performed by the EM or K-means algorithms in the case of model-based clustering.

Another drawback of the current methods for constructing the similarity $S$, with or without learning, is that the similarity function does not take into account cluster labels. By contrast, in model based clustering algorithms like EM, each cluster is represented by a different generative model. The user is able to control what parts of the individual cluster models are shared between clusters. For example, for the well-known mixture of Gaussians model, one can parametrize the cluster covariance matrices in different ways, obtaining, say a common covariance matrix for all clusters, spherical but different covariance matrices, etc. [Banfield and Raftery, 1993]. We would like to endow spectral clustering with a similar flexibility. This can be useful in situations when the combination of features that best characterizes a cluster is different for each cluster. For example, in segmenting natural images, it is often found that the cues that best segment different regions of the image are not the same. Figure 1, illustrates this: edges can be used to segment the snow and some parts of the water, but they should not be used as segment separators for the reeds in the foreground, nor for the lake. In this paper we show that unsupervised learning offers a natural way of learning cluster-dependent similarity functions.

We start by introducing notation and some basic facts in section 2, we define the learning problem in section 3 and introduce the new algorithm in 4. Section 5 presents further details about the algorithm implementation. Experimental results are in section 6 and 7 concludes the paper.

## 2 Spectral clustering – notation and background

In spectral clustering, the data is a set of *similarities* $S_{ij}$, satisfying $S_{ij} = S_{ji} \geq 0$, between pairs of points

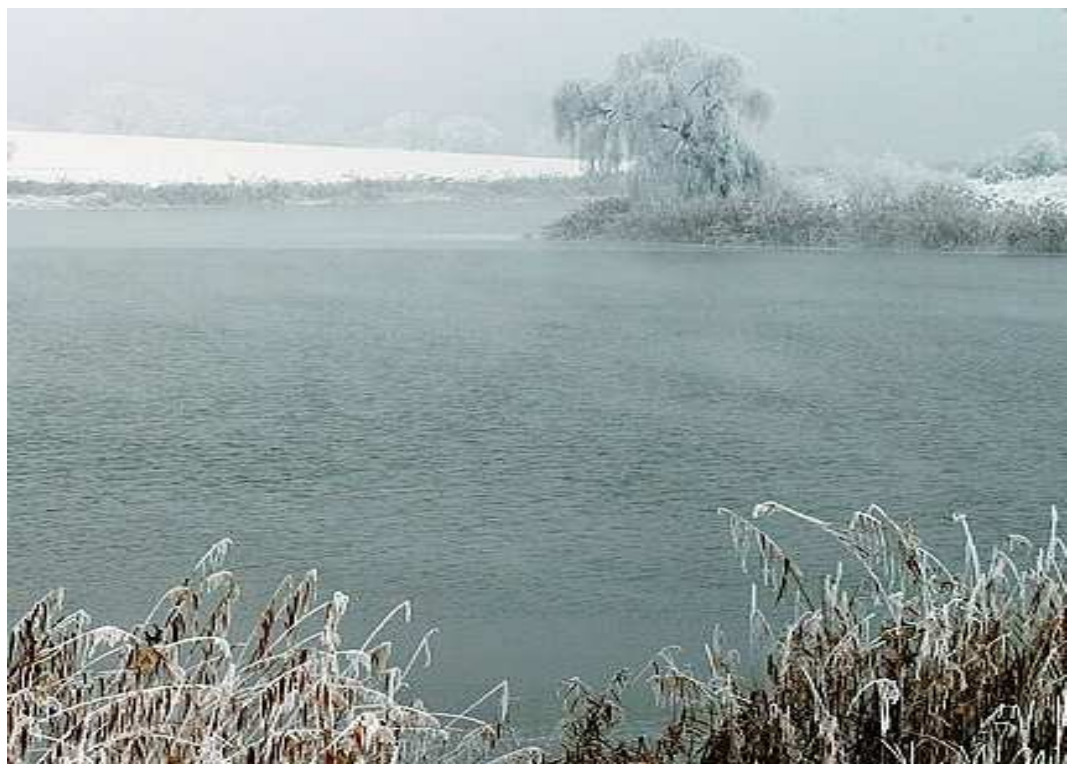

Figure 1: Segmenting pictures is an example where learning parameter which are dependent on the clustering could be helpful.

$i, j$ in a set $V$, $|V| = n$. The matrix $S = [S_{ij}]_{i,j \in V}$ is called the *similarity matrix*. We denote by

$$D_i \equiv Vol\,\{i\} = \sum_{j \in V} S_{ij} \quad (1)$$

the *volume* of node $i \in V$ and by $D$ a diagonal matrix formed with $D_i, i \in V$. The volume of a set $A \subseteq V$ is $Vol\ A = \sum_{i \in A} D_i$. W.l.o.g we assume that no node has volume 0.

**The random walks view** Many properties of spectral clustering are elegantly expressed in terms of the stochastic *transition matrix* $P$ obtained by normalizing the rows of $S$ to sum to 1.

$$P = D^{-1}S \quad \text{or} \quad P_{ij} = S_{ij}/D_i \quad (2)$$

This matrix can be viewed as defining a Markov random walk over $V$, $P_{ij}$ being the *transition probability* $Pr[i \to j|i]$. The eigenvalues of $P$ are $1 = \lambda_1 \geq \lambda_2 \geq \ldots \geq \lambda_n \geq -1$ and the corresponding eigenvectors are $v^1, \ldots v^n$. Note that because $S = DP$ is symmetric, the eigenvalues of $P$ are real and the eigenvectors linearly independent. Define $[\pi_i]_{i \in V}$ by $\pi_i = D_i/Vol\ V$ and $\pi_A = Vol\ A/VolV$ the probability of $A \subseteq V$ under $\pi$.

**The MNCut criterion** A clustering $\mathcal{C} = \{C_1, \ldots, C_K\}$ is defined as a partition of the set $V$ into the disjoint nonempty sets $C_1, \ldots, C_K$. The *multiway normalized cut (MNCut)* clustering criterion [Meilă, 2002, Yu and Shi, 2003]

$$MNCut(\mathcal{C}) = \sum_{k=1}^{K} \sum_{k' \neq k} \frac{Cut(C_k, C_{k'})}{Vol\ C_k} \quad (3)$$

where

$$Cut(A, B) = \sum_{i \in A} \sum_{j \in B} S_{ij} \quad (4)$$

The definition of *MNCut* can be motivated by the Markov random walk view. Let the transition probability between two sets $A, B \subseteq V$ be denoted by $P_{AB} = Pr[A \to B|A]$. Then

$$P_{AB} = \frac{\sum_{i \in A, j \in B} \pi_i P_{ij}}{\pi_A} = \frac{Cut(A, B)}{Vol\ A} \quad (5)$$

It follows that the multiway normalized cut represents the sum of the "out-of-cluster" transition probabilities at the cluster level.

$$MNCut(\mathcal{C}) = \sum_{k=1}^{K} \sum_{k \neq k'} P_{C_k C'_k} = K - \sum_{k=1}^{K} P_{C_k C_k} \quad (6)$$

If $MNCut(\mathcal{C})$ is small for a certain partition $\mathcal{C}$, then the probabilities of evading $C_k$, once the walk is in it, is small.

In [Meilă, 2002] it is shown that the $MNCut(\mathcal{C})$ for any clustering $\mathcal{C}$ is lower bounded by a function of the number of clusters $K = |\mathcal{C}|$ and of the eigenvalues of $P$:

$$MNCut(\mathcal{C}) \geq K - \sum_{k=1}^{K} \lambda_k(P) \quad (7)$$

We call the non-negative difference between the *MNCut* and its lower bound the *gap*:

$$gap_P(\mathcal{C}) = MNCut(\mathcal{C}) - K + \sum_{k=1}^{K} \lambda_k(P) \quad (8)$$

One can show [Meilă, 2002] that the gap is 0 iff $P$ has *piecewise constant eigenvectors (PCE)* $v^1, \ldots v^K$ w.r.t $\mathcal{C}$, that is $v_i^k = v_j^k$ for all $k \leq K$ whenever $i, j$ are in the same cluster.

**Spectral clustering algorithms.** Many spectral clustering algorithms have been formulated. Here we are interested in algorithms that minimize the *MNCut* like the algorithms in [Shi and Malik, 2000, Ng et al., 2002, Meilă and Shi, 2001b]. We use the last one in our experiments and therefore we summarize it here.

**Algorithm** CLUSTER-SPECTRAL
**Input:** Similarity matrix $S$, number of clusters $K$

- Compute transition matrix $P$ by (2)
- Compute $v^1, \ldots v^K$ the eigenvectors corresponding to the of $K$ largerst eigenvalues of $P$
- Clusters the rows of $V = [v^1, \ldots v^K]$ as points in $R^K$ by using e.g K-means.

**Output** clustering $\mathcal{C}$

It can be shown [Meila and Xu, 2003] that when $S$ contains a clustering with a small gap, then the above algorithm will minimize the *MNCut*.

## 3 The learning problem

We assume that we have a data set of size $n$, for which we want to find the best clustering $\mathcal{C}^*$. For each pair of data points $i, j$ in the data set we measure a set of features. The $F$-dimensional vector of features is denoted by

$$x_{ij} = [x_{ij,1}\ x_{ij,2}\ \dots x_{ij,F}]^T \quad (9)$$

The features are *symmetric*, that is $x_{ij} = x_{ji}$ for all $i, j$. We also assume for now that the features are non-negative $x_{ij,f} \geq 0,\ f = 1, \dots F$ and that an increase in $x_{ij,f}$ represents a decrease in the similarity between $i$ and $j$. One can think of the data points as vectors in some $F$-dimensional space, with the features representing distances between points along the $F$ coordinate axes. Our formulation however is significantly more general, in that it accommodates dissimilarity features that do not come from a Euclidean space representation of the data.

The *similarity* is a function $S(x, c, c'; \theta)$ that maps a feature vector $x \in R^F$ and the cluster labels $c, c'$ corresponding to the two points being compared into a non-negative scalar similarity (e.g the well-known Gaussian kernel $S(x; \theta) = e^{-\theta^T x}$, with $\theta \in R^F$). It is assumed that $S$ is symmetric in $c, c'$, i.e $S(., c, c'; \theta) = S(., c', c; \theta)$, and (almost everywhere) differentiable w.r.t the vector of parameters $\theta$. In section 5 we show some simple ways to define cluster-dependent similarity functions.

For a given clustering $\mathcal{C}$, denote by $\mathcal{C}(i) \in \{1, \dots K\}$ the cluster of point $i$ in $\mathcal{C}$. Let

$$S_{ij} = S(x_{ij}, \mathcal{C}(i), \mathcal{C}(j); \theta) \quad \text{for } i, j = 1, \dots n \quad (10)$$

$$\mathbf{x} = [x_{ij}]_{i,j=1,\dots n} \in R^{n \times n \times F} \quad (11)$$

$$S(\theta) \equiv S(\mathbf{x}, \mathcal{C}, \mathcal{C}; \theta) \equiv [S_{ij}]_{i,j=1,\dots n} \in R^{n \times n} \quad (12)$$

Here, the letter $S$ denotes both the similarity function $S(x, c, c'; \theta)$, and the *similarity matrix* $S(\theta)$ for fixed data set $\mathbf{x}$ , clustering $\mathcal{C}$ and parameter vector $\theta$. The matrices obtained from $S(\theta)$ by (1,2) are respectively denoted $D(\theta)$, $P(\theta)$.

The task is to find a set of parameters $\theta$ and a clustering $\mathcal{C}$, such that $\mathcal{C}$ is a "good" clustering for the similarity matrix $S(\mathbf{x}, \mathcal{C}, \mathcal{C}; \theta)$. Note the circularity of the definition, and the fact that the parameters alone do not, in general, uniquely define the clustering $\mathcal{C}$. That in fact the clusters are needed along with the parameters to make the similarity matrix in order to find a clustering. In this respect, our setting differs from both standard spectral clustering and model based clustering, but it does not pose a real problem in an iterative unsupervised setting like the one we introduce in the next section. We assume that the number of clusters $K = |\mathcal{C}|$ is given.

## 4 Algorithm

The algorithm idea is simple for someone familiar with alternate minimization algorithms of the EM type. We assume that we have available a black box spectral clustering algorithm $\mathcal{A}$ that (approximately) minimizes the *MNCut*, like [Meilă and Shi, 2001b]. To measure the quality of a clustering we use the gap defined in (8).

For a given matrix $S$, a clustering that minimizes the gap also minimizes the *MNCut*, so the criteria are equivalent. However, if one *learns* $S$, then the criteria are not equivalent: obtaining a small *MNCut* implies that the off-diagonal blocks of $S$ are nearly 0, while a small gap does not carry such an implication. Hence, the gap puts fewer constraints on $\theta$ while still being an indicator of a good clustering, and we prefer it as a criterion of clustering quality.

Now we can give a first outline of the algorithm

**Algorithm** UNSUPERVISED-SPECTRAL.0

**Inputs:** Data $\mathbf{x}$, initial clustering $\mathcal{C}^0$, initial parameters $\theta^0$, clustering algorithm $\mathcal{A}$.

- Initialize $\theta \leftarrow \theta^0$, $\mathcal{C} \leftarrow \mathcal{C}^0$
- Iterate until convergence:

  **C step:** "Improve clustering"
  Compute $S(\theta) \equiv S(\mathbf{x}, \mathcal{C}, \mathcal{C}; \theta)$. Call $\mathcal{A}$ to obtain a new clustering $\mathcal{C}$.

  **S step:** "Improve similarity"
  Adapt $\theta$ by e.g gradient descent to decrease $gap_{P(\theta)}(\mathcal{C})$

**Output:** $\theta$, $\mathcal{C}$

The algorithm alternates between two steps: in the **C** step, a "best" clustering is found for the given similarities; the **S** step is a supervised learning step, with the target clustering obtained in the **C** step. We have presented this naive algorithm for the sake of simplicity, but the careful reader will have already spotted some of its weak points. We discuss and fix them now.

It has been shown in [Meilă et al., 2005] that supervised learning algorithms that minimize the gap on the training data are apt to be overfitting. In [Meilă et al., 2005] it was also shown that if a cluser-ing which has a small gap and a large $\Delta_k$, all other clusterings with a small gap will be close to it. Therefore the authors introduce a regularized learning criterion, that balances the quality term $gap_{P(\theta)}(\mathcal{C})$ with a stability term represented by the (squared) eigengap $\Delta_K(\theta) = \lambda_K(P(\theta)) - \lambda_{K+1}(P(\theta))$. We adopt this idea here. Hence, in the **S** step, we will decrease (or

minimize) the criterion

$$f_\alpha(\theta, \mathcal{C}) = gap_{P(\theta)}(\mathcal{C}) - \alpha\Delta_K^2(P(\theta)) \quad (13)$$

with $\alpha$ a regularization parameter. We choose the squared form because empirically we find it to give better numerical behavior [Meilă et al., 2005].

A problem with the **C** step is that spectral clustering algorithms like [Meilă and Shi, 2001b, Ng et al., 2002, Shi and Malik, 2000] only minimize the *MNCut* in a neighborhood of a $P$ with piecewise constant eigenvectors. For an arbitrary $S$ (or $P$), there are no theoretical guarantees that the *MNCut* is optimized. Moreover, the final clustering is usually obtained by the K-means algorithm. There are no guarantees of global optimization for K-means, and its strong dependence on initialization is well documented. To counteract the variability of the algorithm $\mathcal{A}$ we adopt two remedies. The first one, used before by various authors, is to use K-means with multiple initializations, including orthogonal centers initializations that are particularly well suited for the spectral clustering setup [Ng et al., 2002, Verma and Meilă, 2003]. A set of candidate clusterings $\mathcal{M}$ is obtained. From it the "best" clustering is usually chosen to minimize the K-means distortion. Here we compare the clusterings by their *MNCut*. Comparing by *MNCut* (or equivalently by the gap) is practically very similar to comparing by distortion, as the two measures can be shown to be related [Meila and Xu, 2003]. Second, recalling that the similarity matrix was obtained using the clustering $\mathcal{C}$ from the previous iteration, we add $\mathcal{C}$ to the set of candidates $\mathcal{M}$. We now obtain

**Algorithm** UNSUPERVISED-SPECTRAL

**Inputs:** Data $\mathbf{x}$, initial clustering and parameters $\mathcal{C}^0$, $\theta^0$, regularization constant $\alpha$, clustering algorithm $\mathcal{A}$

- Initialize $\theta \leftarrow \theta^0$, $\mathcal{C} \leftarrow \mathcal{C}^0$
- Iterate until convergence:
  - **C** Compute $S(\theta) \equiv S(\mathbf{x}, \mathcal{C}, \mathcal{C}; \theta)$
    Call $\mathcal{A}$ with different random bits to obtain a set of clusterings $\mathcal{M}$.
    $\tilde{\mathcal{C}} = \text{argmin}_{\mathcal{C}' \in \mathcal{M}} gap(\mathcal{C}')$
    If $gap(\tilde{\mathcal{C}}) < gap(\mathcal{C})$ then $\mathcal{C} \leftarrow \tilde{\mathcal{C}}$
  - **S** Adapt $\theta$ to decrease $f_\alpha(\theta, \mathcal{C})$.

**Output:** $\theta$, $\mathcal{C}$

**Proposition 1** *The algorithm* UNSUPERVISED-SPECTRAL *converges for any initial values* $(\theta^0, \mathcal{C}^0)$ *and any* $\alpha$.

**Proof** (Sketch) Every time the **C** step changes $\mathcal{C}$, $f_\alpha$ is strictly decreased. The same happens whenever $\theta$ is changed in the **S** step[1]. As the gap is non-negative and the eigenvalues of $P$ lie in $[-1, 1]$, $f_\alpha(\theta, \mathcal{C}) \geq -4\alpha$ for any $\theta, \mathcal{C}$. Hence, by a Liapounov function argument, the sequence of $(\theta, \mathcal{C})$ values will converge to a local minimum of $f_\alpha$[2]. ■

In our experiments we use gradient descent with line search by the Armijo rule for the **S** step, and the algorithm described in section 2 for the **C** step.

The problem is generally non-convex, so the limit point will depend on the initial values. Other factors, like the number of optimization steps in the **S** step and the nature of the randomization in the **C** step can also influence the value of the algorithm's fixed point.

## 5 Practical considerations

**The similarity function** $S$**.** A simple way to construct a label dependent similarity function is to have a different set of parameters for each cluster or cluster pair. For example, let $\theta = \{\theta^{c,c'} \in R^F,\ c, c' = 1, \ldots K\}$ and

$$S(x_{ij}, c(i), c(j); \theta) = \exp\left[\sum_{f=1}^{F} \theta_f^{c(i)c(j)} x_{ij}^f\right] \quad (14)$$

This amounts to having $FK^2$ free parameters. More compact parametrization $\theta = \{\theta^c \in R^F,\ c = 1, \ldots K\}$, involving one parameter vector for each cluster, are

$$S(x_{ij}, c(i), c(j); \theta) = \exp\left[\sum_{f=1}^{F} \theta_f^{c(i)} \theta_f^{c(j)} x_{ij}^f\right] \quad (15)$$

$$S(x_{ij}, c(i), c(j); \theta) = \exp\left[\sum_{f=1}^{F} (\theta_f^{c(i)} + \theta_f^{c(j)}) x_{ij}^f\right] \quad (16)$$

The above expression model a "soft" AND and respectively a "soft" OR for the relevance of the $f$-th feature. In our experiments we use the product form (15). Generalizations to any non-negative function or kernel, e.g $K(\theta^{c,c'}, x)$ are immediate.

To avoid as many as possible of the shallow local minima we adopt several heuristic techniques with a regularizing or smoothing effect.

**The amount of optimization in the S step** At the beginning of the algorithm, when $\mathcal{C}$ is likely not a reliable clustering, it is reasonable to only change $\theta$ by small amounts before re-evaluating $\mathcal{C}$. The opposite is true near convergence, when we can take many steps in $\theta$ before changing $\mathcal{C}$ (this will also save time, as $\mathcal{C}$ becomes more stable the **C** step is less likely to change it even if performed). Hence, we decide to terminate the **S** step after an iteration on $\theta$ with probability $p_{reclust}$. After every learning step with probability $p_{reclust}$ the spectral clustering algorithm is performed with the current parameter values. If at least one new clustering was found, we have evidence that our clusterings are not stable so we stop learning with the current clusterings and perform our **C** step to obtain a the set of new clusterings in order to learn. If no new clusterings are found this provides evidence that our clusterings are stable. When this happens $p_{reclust}$ is reduced to increase the efficiency of the algorithm and learning with the same clustering(s) is continued.

[1]Any optimization algorithm can easily be modified to ensure this, by simply stopping whenever it cannot decrease the objective.

[2]To rigorously define a local minimum, one must define a topology on the space of clusterings $\mathcal{C}$. Noting that this space is discrete and finite, we skip further discussion of this point.

**Regularizing the C step** We also make the **C** step less committal in the early stages of the algorithm, by producing more than one target clustering. Adapting the parameters $\theta$ is done by minimizing the weighted sum

$$\sum_{\mathcal{C}_i \in \mathcal{T}^{(q)}} w_i f_\alpha(\mathcal{C}_i, \theta) \tag{17}$$

with $\{\mathcal{C}_i\} = \mathcal{T}^{(q)}$ the set of target clusterings produced in the $q$-th **C** step. Since the only part of $f_\alpha$ that depends on $\mathcal{C}$ is the *MNCut*, we will be effectively replacing the gradient w.r.t to the *MNCut* with a weighted sum of these gradients, one for each $\mathcal{C}_i$ in the target set. The modification of the **C** step is as follows

**Algorithm** C-MULTIPLE-TARGETS

**Inputs:** Parameters $\theta^{(q-1)}$, target clusterings $\mathcal{T}^{(q-1)}$.

1. Compute $S(\theta) \equiv S(\mathbf{x}, \mathcal{C}, \mathcal{C}; \theta)$ where $\mathcal{C}$ is the lowest *MNCut* clustering in $\mathcal{T}^{(q-1)}$.
2. Call $\mathcal{A}$ with different random bits to obtain a set of clusterings $\mathcal{M}$.
3. $\mathcal{M} \leftarrow \mathcal{T}^{(q-1)} \cup \mathcal{M}$.
4. Let $r_i = \frac{MNCut(\mathcal{C}_i)}{\min_{\mathcal{C}' \in \mathcal{M}} MNCut(\mathcal{C}')}$ for $\mathcal{C}_i \in \mathcal{M}$.
5. $\mathcal{T}^{(q)} \leftarrow \{\mathcal{C}_i \in \mathcal{M} \mid r_i < e^{[1-\Delta_K(P(\theta))]^2}\}$
6. $w_i = \frac{e^{-MNCut(\mathcal{C}_i)}}{\sum_{i'} e^{-MNCut(\mathcal{C}_{i'})}}$ for $\mathcal{C}_i \in \mathcal{T}^{(q)}$.

**Output:** $\mathcal{T}^{(q)}$, $\{w_i, \mathcal{C}_i \in \mathcal{T}^{(q)}\}$.

In step 6 of the above algorithm, the cutoff function $e^{(1-\Delta_K)^2}$ will become more restrictive as the eigengap increases; also, if a clear winner emerges in the set $\mathcal{M}$, the size of the target set will decrease. The goal, as stated before, is to average the gradient over several clusterings in the early stages of the algorithm, but to limit the target set to a single clustering in the later stages. The weights $w_i$ represent the mixing weights of the gradients w.r.t to the normalized cuts of the clusterings. By making this modification, we cannot guarantee the convergence of the algorithm in all cases anymore. However, in the experiments we have found this not to be a problem.

**Selecting the PCE's.** By studying the eigenvectors of $P(\theta)$ in the earlier stages of learning, we have found that the eigenvectors most indicative of a good clustering, or simply the ones "more piecewise constant" are often interspersed with spurious eigenvectors. Thus, selecting the first $K$ eigenvectors will often lead into local optima characterized by poor clusterings. This is why we have introduced an additional modification to the **C** step, which consists of actively selecting the $K$ eigenvectors that will be used to cluster the data out of a larger set of $K'$ vectors. We do so by defining an *index of PC-ness* denoted $pc(v)$ for every eigenvector $v$ but the constant one. The index is a measure of the "smoothed entropy" of the vector $v$ seen as a set of $n$ points. The full description of the index is omitted here for lack of space. This method is particularly effective in problems where there are naturally many large spurious eigenvalues, like for instance problems with "non-compact" clusters, where the intra-cluster similarities have high variance and there are large numbers of small values. In the experiments, we take $K' = \max(2K, 10)$.

Selecting other than the first $K$ eigenvectors to cluster by entails some changes to the criterion we optimize. Let $v^{i_1}, v^{i_2}, \ldots v^{i_K}$ be the chosen eigenvectors, with $1 = i_1 > i_2 > \ldots > i_K$ and $i_0$ be the index of the first eigenvector not in $\{v^{i_1}, v^{i_2}, \ldots v^{i_K}\}$. Then, the new criterion is

$$\tilde{f}_\alpha(\theta, \mathcal{C}) \;=\; [MNCut_{P(\theta)}(\mathcal{C}) - K + \sum_{k=1}^{K} \lambda_{i_k}] - \alpha(\lambda_{i_K} - \lambda_{i_0}) \tag{18}$$

The change in the gap term reflects the fact that we want the most PC eigenvectors to be the ones influencing $\theta$ (recall that the derivative of $\lambda$ w.r.t the matrix elements, when it exists, is given by $vv^T$); the change in the eigengap term forces the relevant subspace $(v^{i_1}, v^{i_2}, \ldots v^{i_K})$ to become a stable principal subspace by pushing upward the corresponding eigenvalues. Note that with these changes neither the gap nor the eigengap are guaranteed to be positive.

## 6 Experiments

The results of experiments run on three data sets are presented here. algorithm. The first data set is an artificial data set, *Gaussians*, the second is the *Dermatology* dataset obtain from the UCI machine learning repository and the third is an image, *Cow*, to be segmented. A brief summary of the data sets used in the experiments are listed in Table 1. The number of clusters, $K$ is known for the *Gaussian* data set and the *Dermatology* data. For the *Cow* image $K$ is unknown, thus two different values are tried and the results for both are shown. The similarity function used is defined as in (15) for the label-dependent parameters or simply as $S = \exp(\theta^T x)$ for the label-independent parameters. The parameters are always initialized with equal values, $p_{reclust}$ is intialized to 0.8 and the maximum number of target clusterings kept is six.

The true clusterings are known in the *Gaussain* and *Dermatolory* case so the classification error is reported. The classification error is defined as the minimum proportion of misclassified data points over all permutations of the esimated cluster labels. Permuting over the estimated cluster labels finds the best match up between the true assignment and the esimated clustering. The *Cow* image is unsupervised, thus classification error can not be reported, and just the segmentation results are reported in figure 4.

Table 1:

| **Data set** | $K$ | $n$ | # features |
|---|---|---|---|
| 4 Gaussians | 4 | 350-500 | 4-10 |
| Dermatology | 6 | 358 | 34 |
| Cow | 4,5 | 1024 | 7 |

**Gaussian** This artificial data set contains four separated two-dimensional Gaussian clusters with varying cluster size. Noisy dimensions are added to the meaningful features, the horizontal and vertical coordinates, to test the algorithms ability to learn which features are important in determining the cluster assignments. We learn a single set of parameters $\theta$ for all clusters and Table 2 reports the results of these experiments. All parameters are intialized to a uniform value, and it is these intial values which are used as weights for the column titled "Before Learning".

For all levels of noisy dimensions added to the meaninful features the clustering error is lower after the learning process. In addition to a lower clustering error, the gap is decreased, which is a measure of cluster quality and the eigengap is increased. After the learning process the parameters associated with the noisy features are zero, which is desirable since they provide no information toward the correct clustering in the data.

**Dermatology** This data set was obtained from the UCI Machine Learning Repository [Guvenir and Ilter, 1998]; it contains information on patients with Eryhemato-Squamous disease. There are thirty-four features in this dataset, demographic information, family history and various skin measurements are included. Eight of the original 366 patients are excluded due to missing information, leaving 358 patients. The intial values are once again uniform for all features and are used as the "before learning" parameter values. Two sets of weights are learned for this data set, the first set is the label-independent parameters which reduce the clustering error to 16%. The second set of parameters are cluster dependent parameters as described in (15). Table 3 reports the results of these learning experiments for both sets of parameters. The clustering error is decreased substantially, along with the gap for the label independent parameters. There is variability in the parameter values for each feature between clusters, suggesting that label-dependent weights may prove helpful. Figure 2 plots the feature weights, both the label independent as well as the label dependent for each cluster. The label-indepedent parameters are quite a bit smaller than the cluster-dependent parameters. While there is a disctinct pattern over all the clusters, there is variation between clusters in the cluster-dependent parameters, especially for the last feature which corresponds to a patients age. formulation choice for the similarity function is not general enough, explaining why even though the weights themselves show that parameters specific to each cluster are useful, the difference in classification error is small.

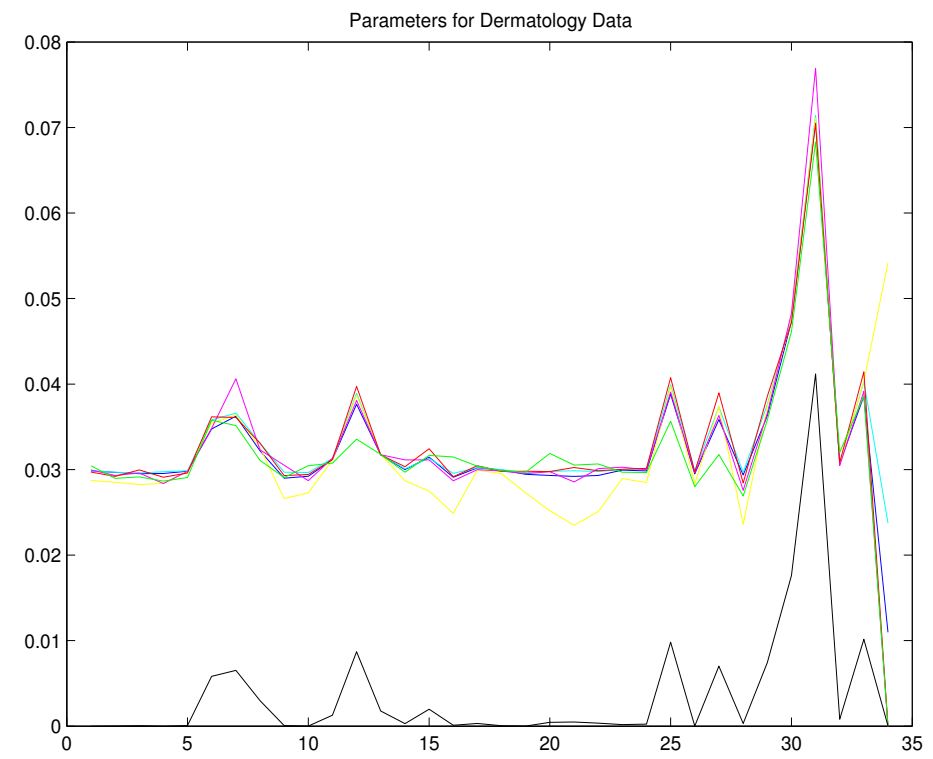

Figure 2: **Dermatology Weights** Learned parameters for *Dermatology* data. The black line is the values for the label-independent parameters. The colored lines each present the cluster-dependent parameter values for a cluster.

**Cow** The final experiment involves a thirty-two by thirty-two color image of an animated cow shown in Figure 3. The features are pairwise between pixels, with three corresponding to color and four features corresponding to texture filters. In this experiment, K is unknown thus the parameters are learned assum-ing $K = 4$ and $K$=5 and the segmentation evaluated by eye. For both $K$=4 and 5, the segmentation before learning is haphazard. The segmentation after learning assuming $K = 4$ looks good. There are a few noticeable errors but for the most part the colorings, which correspond to clusters look appropriate. For $K$=5 the segmentation after learning looks okay, yet there appear to be more errors than $K$=4 suggesting the fifth cluster is unnecessary.

Table 2: **Gaussian Clusters** The results presented here correspond to an experiment run on four bivariate clusters with different means, variance and cluster sizes. There are two meaningful features for the clustering the horizontal and vertical coordinates; in addition to these two noisy features are added which provide no information about the proper clustering. "# Noisy Dims" indicates the number of non-meaninful features added. "Before Learning" presents the results from using uniform parameter values for all features.

| | | After Learning | | | | Before Learning | | | |
|---|---|---|---|---|---|---|---|---|---|
| # Noisy Dims | $n$ | CE | gap | eigengap | MNCut | CE | gap | eigengap | MNCut |
| 2 | 350 | 0.0257 | 0.8698 | 0.3376 | 1.3302 | 0.0514 | 0.8869 | 0.0565 | 1.4443 |
| 4 | 500 | 0.0140 | 0.8884 | 0.3268 | 0.1976 | 0.4380 | 0.9559 | 0.0364 | 1.7519 |
| 8 | 500 | 0.0160 | 0.7058 | 0.3487 | 0.9271 | 0.6920 | 1.4578 | 0.0026 | 1.5925 |

Table 3: **Dermatology** Clustering results for *Dermatology* data set learning experiments.

| Parameters | CE | gap | eigengap | MNCut |
|---|---|---|---|---|
| Uniform $\theta$) | 0.4385 | 0.6394 | 0.0232 | 4.23 |
| Learned Label-Independent $\theta$ | 0.1508 | 0.0014 | 7.15e-9 | 4.95 |
| Learned Label-Dependent $\theta$ | 0.1704 | 0.0085 | 9.0e-3 | 5.0 |

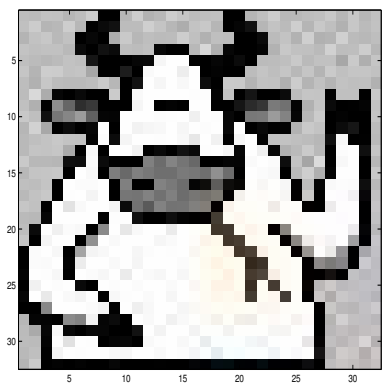

Figure 3: **Cow** Animated cow image used for *Cow* experiment.

## 7 Discussion

We have introduced unsupervised learning for spectral clustering in a form reminiscent to model-based clustering by mixture models. The clustering and parameters of the model are learned simultaneously. The algorithm, in the form given in section 4, is converging to a local minimum of the cost function $f_\alpha(\theta, \mathcal{C})$ which represents the sum of the gap and the regularization term eigengap based on the eigengap. Currently we assume that $K$ and $\alpha$ are known; these are strong assumptions rarely warranted in practice. Removing them is our first priority for future research. In this aspect, unsupervised spectral learning is at a disadvantage w.r.t model based clustering, where the presence of a probabilistic model allows one to apply statistical model selection methods to the task. Another similarity of the two methodologies is the presence of local minima. Our experiments have shown that this problem exists in practice, but that in spite of it our algorithms are still capable to find meaningful clusterings.

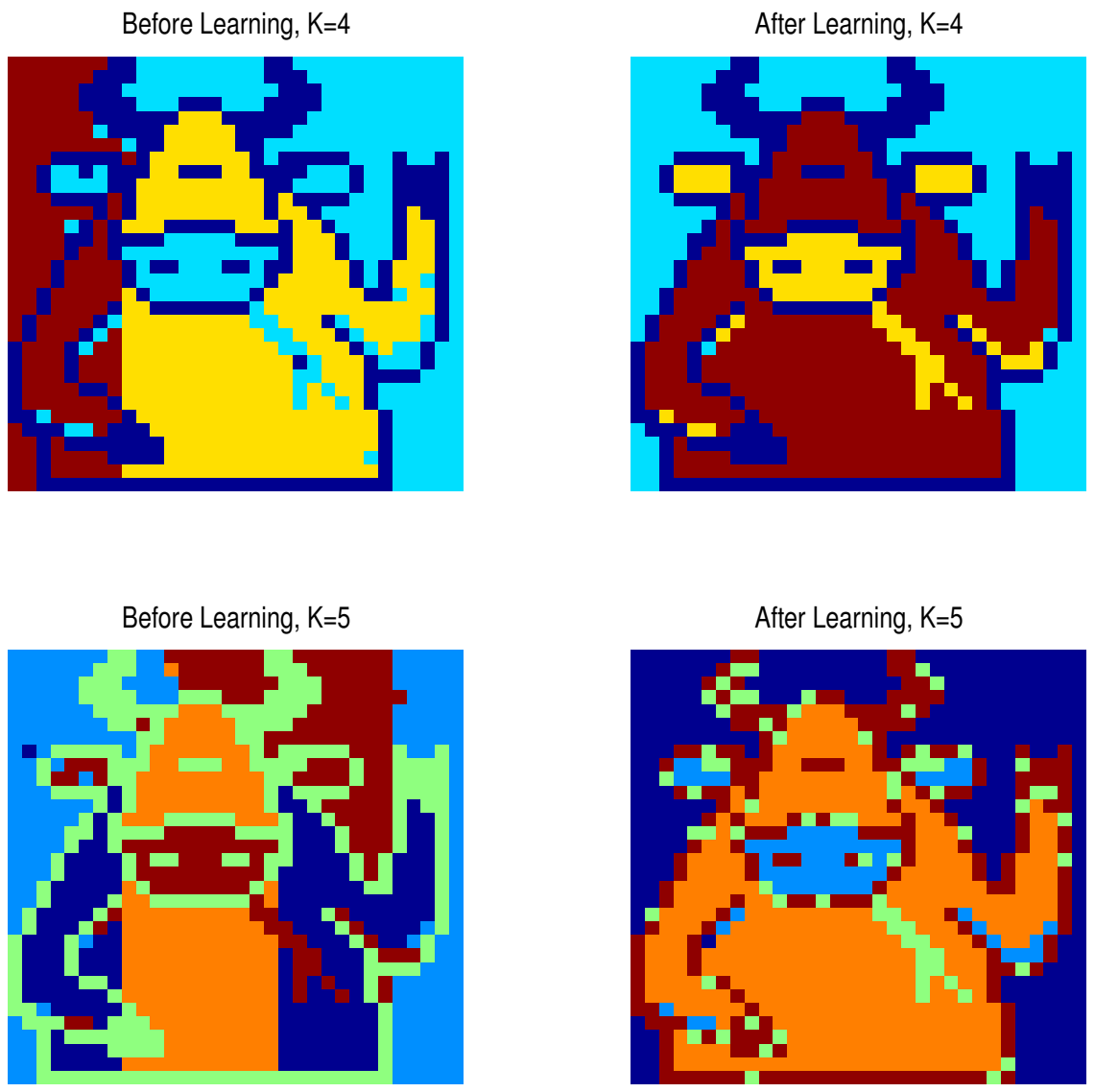

Figure 4: **Cow** Learning parameters for segmenting an image of a cow. The image is 32 by 32 pixels, and the features are color and texture.

Finally, we shall recall the connection between the black box clustering algorithm $\mathcal{A}$ and the (unregularized) cost function $f_0(\theta, \mathcal{C}) = gap_{P(\theta)}(\mathcal{C})$. The framework we propose immediately extends to other spectral clustering algorithms that can be shown to (locally) minimize a given cost function, like for instance the algorithm of [Bach and Jordan, 2003].

**Acknowledgements**

This work was partially supported by NSF VIGRE grant 621261 and NSF ITR grant 0313339.

## References

[Bach and Jordan, 2003] Bach, F. and Jordan, M. I. (2003). Learning spectral clustering. Technical Report CSD-03-1249, University of California, Berkeley.

[Bach and Jordan, 2004] Bach, F. and Jordan, M. I. (2004). Learning spectral clustering. In Thrun, S. and Saul, L., editors, *Advances in Neural Information Processing Systems 16,*, Cambridge, MA. MIT Press.

[Banfield and Raftery, 1993] Banfield, J. D. and Raftery, A. E. (1993). Model-based gaussian and non-gaussian clustering. *Biometrics*, 49:803–821.

[Cour et al., 2005] Cour, T., Gogin, N., and Shi, J. (2005). Learning spectral graph segmentation. In Cowell, R. and Ghahramani, Z., editors, *Artificial Intelligence and Statistics Workshop (AISTATS05).*

[Guvenir and Ilter, 1998] Guvenir, H. and Ilter, N. (1998). UCI repository of machine learning databases.

[Meila and Xu, 2003] Meila, M. and Xu, L. (2003). Multiway cuts and spectral clustering. Technical Report 442, University of Washington. `www.stat.washington.edu/mmp/Papers/nips03-multicut-tr.ps`.

[Meilă, 2002] Meilă, M. (2002). The multicut lemma. Technical Report 417, University of Washington.

[Meilă and Shi, 2001a] Meilă, M. and Shi, J. (2001a). Learning segmentation by random walks. In Leen, T. K., Dietterich, T. G., and Tresp, V., editors, *Advances in Neural Information Processing Systems*, volume 13, pages 873–879, Cambridge, MA. MIT Press.

[Meilă and Shi, 2001b] Meilă, M. and Shi, J. (2001b). A random walks view of spectral segmentation. In Jaakkola, T. and Richardson, T., editors, *Artificial Intelligence and Statistics AISTATS.*

[Meilă et al., 2005] Meilă, M., Shortreed, S., and Xu, L. (2005). Regularized spectral learning. In Cowell, R. and Ghahramani, Z., editors, *Proceedings of the Artificial Intelligence and Statistics Workshop(AISTATS 05).*

[Ng et al., 2002] Ng, A. Y., Jordan, M. I., and Weiss, Y. (2002). On spectral clustering: Analysis and an algorithm. In Dietterich, T. G., Becker, S., and Ghahramani, Z., editors, *Advances in Neural Information Processing Systems 14*, Cambridge, MA. MIT Press.

[Shi and Malik, 2000] Shi, J. and Malik, J. (2000). Normalized cuts and image segmentation. *PAMI.*

[Verma and Meilă, 2003] Verma, D. and Meilă, M. (2003). A comparison of spectral clustering algorithms. TR 03-05-01, University of Washington. (sumitted).

[Yu and Shi, 2003] Yu, S. X. and Shi, J. (2003). Multiclass spectral clustering. In *International Conference on Computer Vision.*